\documentclass[final,1p,times,twocolumn]{elsarticle}
\usepackage[utf8]{inputenc}

\usepackage{amsmath}
\usepackage{algorithm}
\usepackage[noend]{algpseudocode}
\usepackage{subfigure,underoverlap}
\usepackage{multirow}
\usepackage{color}
\usepackage{cancel}
\usepackage{soul}

\title{An Unsupervised Multivariate Time Series Kernel Approach for Identifying Patients with Surgical Site Infection 
from Blood Samples}

\author[UiT,ML]{Karl Øyvind Mikalsen\corref{cor1} }
\address[UiT]{Dept. of Mathematics and Statistics, UiT The Arctic University of Norway, Tromsø, Norway}
\address[ML]{UiT Machine Learning Group}

\cortext[cor1]{Corresponding author at: Department of Mathematics and Statistics, Faculty of Science and Technology,
UiT – The  Arctic University of Norway,
N-9037 Tromsø, Norway}

\ead{karl.o.mikalsen@uit.no}

\author[ML,spain]{Cristina Soguero-Ruiz}
\address[spain]{Dept. of Signal Theory and Comm., Telematics and Computing, Universidad Rey Juan Carlos, Fuenlabrada, Spain}

\author[MLIFT,ML]{Filippo Maria Bianchi}
\address[MLIFT]{Dept. of Physics and Technology, UiT, Tromsø, Norway}

\author[UNN2,UNN3,UNN4]{Arthur Revhaug}
\address[UNN2]{Dept. of Gastrointestinal Surgery, University Hospital of North Norway (UNN), Tromsø, Norway}
\address[UNN3]{Clinic for Surgery, Cancer and Women's Health, UNN, Tromsø, Norway}
\address[UNN4]{Institute of Clinical Medicine, UiT, Tromsø, Norway}

\author[MLIFT,ML]{Robert Jenssen}

\begin{document}

\begin{frontmatter}
%\maketitle

\begin{abstract}
A large fraction of the electronic health records consists of clinical measurements collected over time, such as blood tests, which provide important information about the health status of a patient. 
These sequences of clinical measurements are naturally represented as time series, characterized by multiple variables and the presence of missing data, which complicate analysis.
In this work, we propose a surgical site infection detection 
framework for patients undergoing colorectal cancer surgery that is completely unsupervised, hence alleviating the problem of getting access to labelled training data. %Missing data  are handled by powerful kernels for computing the similarity between  multivariate time series. 
The framework is based on powerful kernels for multivariate time series that account for missing data when computing similarities.
Our approach show superior performance compared to baselines that have to resort to imputation techniques and performs comparable to a supervised classification baseline.
\end{abstract}


\end{frontmatter}

%********************
% Introduction
%********************

\section{Introduction}
Surgical Site Infection (SSI) is one of the most common types of nosocomial infections~\cite{lewis2013}, representing up to 30\% of hospital-acquired infections~\cite{magill2012prevalence, de2009surgical}.  
SSI can be divided into different types depending on the anatomical location of the infection~\cite{ko2015american}.
\emph{Superficial} infections can be treated with local surgical debridement and antibiotics.
On the other hand, \emph{deep} infections are more complex and require lapratomies and/or percutaneous drainage and intravenous antibiotics.
Recently, SSI risk factors such as advanced age, overweight, smoking, open surgery or disseminated cancer have been reported~\cite{lawson2013risk}.  Depending on the location of the infection, some factors contribute more to increase the risk of SSI.
For example, longer lasting surgeries are associated with deep SSI, whereas a high body mass index is related with superficial SSI~\cite{lawson2013risk,blumetti2007surgical}.

Along with increased mortality rate (3\%), SSI also prolongs hospitalization up to two weeks~\cite{whitehouse2002} and increases the risk of readmission~\cite{shah2017evaluation}.
This, in average, doubles the expenses per patient and increases the chances of readmission, with further additional costs up to 27,000 USD~\cite{whitehouse2002, owens2014surgical}. Hence, a reduction in the number of postoperative complications like SSI will be of great benefit both for the patients and the society.

Many recent studies have focused on the analysis of SSI from blood tests, both before and after surgery~\cite{silvestre2014, soguero2015data,medina2016,angiolini2016}.
The advantage of using blood tests for this purpose is that they are recorded frequently with low burden for the patients and contain much information about their actual health status.
Among blood tests results, the high predictive power of C-reactive protein (CRP) test has been evaluated and emphasized in several works. For example, authors in~\cite{angiolini2016} and~\cite{gans2015diagnostic}, demonstrated the relation between different CRP cutoff values and the risk of SSI  on postoperative days 3 and 4. Others have combined the results from blood tests with other structured and unstructured clinical data to predict SSI complications~\cite{soguero2016support,hu2017strategies,Sanger2016259}.

Since blood samples are collected over time they can naturally be represented as \emph{time series}. 
Such clinical time series data have some special characteristics that distinguish them from time series from other domains. 
One key property is \emph{missing data}, which can occur because of e.g. lack of documentation or lack of collection~\cite{wells2013strategies}. 
Another characteristic is that the time series usually are \emph{multivariate}. For example could the patients be described by the measurements of many different blood tests, such as e.g. hemoglobin, CRP, etc., where many of them exhibit relationships or dependencies, which can be non-trivial (non-linear). 

In order to apply a machine learning algorithm on such time series, one can work directly in the input space using a \emph{similarity measure} that accounts for dependencies in time and among the variables~\cite{mikalsen2017time}. 
An advantage is that a time consuming feature learning process, or a manual feature design process that requires user intervention and domain expertise, can be then avoided. 
On the other hand, a key problem with classical similarity measures is that most of them are not able to cope with missing data in clinical time series data without applying a preprocessing step such as imputation in order to obtain complete data~\cite{mitsa2010temporal}. However, important information about the clinical condition of the patient and the decisions of the caregiver may disappear in such a preprocessing step since the information that some data are missing is lost, and replaced by biased estimates.

In addition to the problem of analyzing multivariate time series (MTS) containing missing data, another key challenge associated with data originating from electronic health records (EHRs) is that getting access to labelled data for training the machine learning models is difficult. 
It is well known that manual annotation of labels, especially in the healthcare domain, is a cumbersome process that could be both time consuming and expensive~\cite{MIKALSEN2017105}, since clinical expertise is needed to create the training sets. 
Indeed, the workload for the clinicians is tremendous already, and with an aging and more diseased population we cannot  expect these tasks to be prioritized in the future. To overcome this annotation problem, recently there have been proposed several different methods within the framework of semi-supervised learning~\cite{chapelle2009semi}, of which the so-called anchor learning is an example~\cite{MIKALSEN2017105,Halpernocw011}. 

In this work, we propose a disease classification framework for time series originating from EHR data such as e.g. blood tests. We address the key challenges described above by taking an unsupervised approach where we utilize powerful kernels for MTS containing missing data. 
These kernels have emerged due to recent advances in time series analysis as new family of methods that can cope with incomplete and multivariate data.  Prominent examples include the \emph{learned pattern similarity} (LPS)~\cite{baydogan2016time} and the  \emph{time series cluster kernel} (TCK)~\cite{mikalsen2017time}.  These methods are able to handle missing data without having to resort to imputation methods. 
Another important property of the LPS and TCK kernels is that they can be trained in an unsupervised way and do not require tuning of critical hyperparameters. 
We take advantage of their robustness by using the kernels as input to an unsupervised \emph{spectral clustering} framework for identifying patients with SSI. 
By doing so, the entire framework is grounded within the theoretically well understood kernel methods. 
Moreover, spectral clustering is considered a state-of-the-art clustering algorithm and has been successfully utilized in many applications~\cite{ng2002spectral,krzakala2013spectral, lokse2017spectral}. 

\begin{figure}[!t]
    \centering
    \includegraphics[trim = 0mm 0mm 0mm 0mm, clip, width=\linewidth]{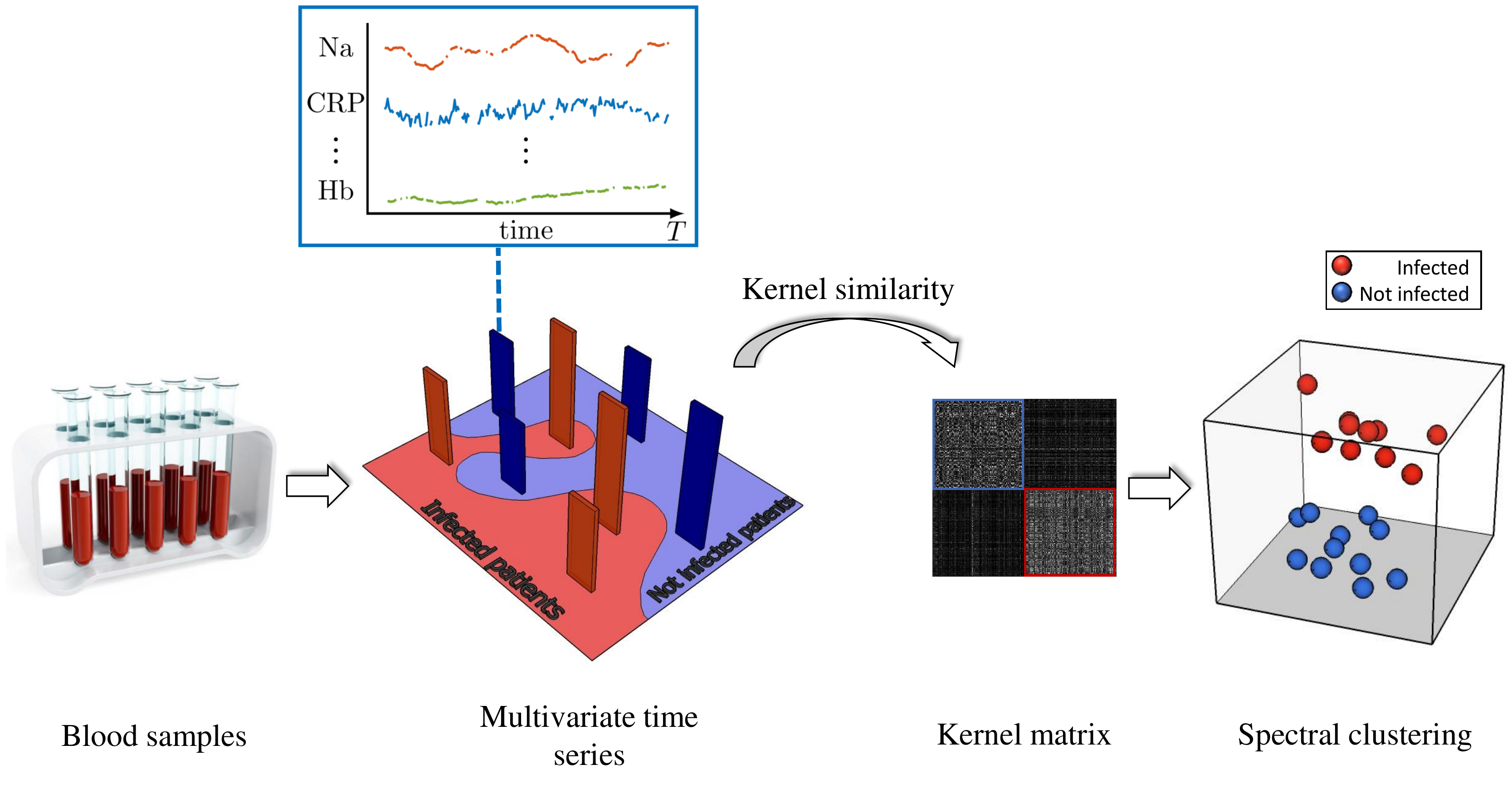}
         \caption{\textit{Overview of the unsupervised kernel approach for predicting postoperative SSI from MTS blood samples.}}
    \label{fig:Scheme}
\end{figure}

The proposed methodology consists of two steps, namely to first compute the kernel and then apply spectral clustering  (see Fig.~\ref{fig:Scheme}). 
We use the proposed framework to identify patients undergoing colorectal cancer surgery with SSI based on only blood samples, and illustrate its power by
comparing to a similar framework
where we use kernels that have to resort to imputation techniques. In addition, we compare to a supervised baseline for detecting SSI.

The rest of the paper is as follows. Section~\ref{Methods} describes materials and methods. 
Results are presented in Section~\ref{Sec:Results}. In Section~\ref{sec: discussion} we provide a discussion, whereas a conclusion is drawn in the last section.

%#####################################
\section{Materials and methods}
\label{Methods}
%#####################################

%#####################################
\subsection{Data description} \label{Database}
%#####################################

Ethics approval for the present study was obtained from the Data Inspectorate and the Ethics Committee at the University Hospital of North Norway (UNN)~\cite{jensen2017analysis, soguero2016predicting}. 
The dataset we consider consists of  7741 patients that underwent a gastrointestinal surgical procedure at UNN in the years 2004–2012. 
The SSI persistent in-hospital morbidity is particularly associated with colorectal cancer surgery~\cite{lawson2013risk,blumetti2007surgical,lawson2013reliability} and therefore patients that did not undergo this type of surgery were excluded, reducing the size of the cohort to 1137 patients.

The International Classification of Diseases (ICD10) or NOMESCO Classification of Surgical Procedures (NCSP) codes related to severe postoperative complications, both superficial and deep infections, were considered to extract a cohort. %\commentF{should not this go before reporting the number of samples in the dataset?} \textbf{what we describe above is not the final cohort}. 
For the purpose of this work and similarly to earlier studies~\cite{LIMON2014127,Gibbons20111,horan_gaynes_martone_jarvis_emori_1992,BERGER2013974,soguero2015data,Sanger2016259}, we do not distinguish between deep and superficial SSI. % \commentR{Do we need to explain why we don't do the distinction?} \textbf{In the Sanger paper they just have a sentence similar to the one we use here, but I added 4 more refs (the same that Sanger used), so now there are 6 refs in total backing up this} .

% A patient without having any of these codes was considered as a control, otherwise, as a case.  We ended up with a total of YYY infected patients (cases) and XXX non-infected ones (control). A demographic description of this cohort is showed in Table~\ref{tab:demographics}.

In collaboration with the clinician (author A. R.), a set of 11 different types of blood tests were defined as clinically relevant  and extracted for all patients from their EHRs, namely, hemoglobine, leukocytes, CRP, potassium, sodium, creatininium, thrombocytes, albumin, carbamide, glucose and amylase. The blood tests can be considered as continuous variables  over time and hence represented as MTS. 
For the purpose of the current analysis, we discretize time and let each time interval be one day. 
However, all 11 blood tests are not available every day for each patient, meaning that the dataset contains missing data. 
We focus on classification of SSI within 20 days after surgery and therefore define the \emph{postoperative window} as the period from postoperative day 1 until day 20. Patients with less than two measurements during the postoperative window are removed from the cohort, which leads to a dataset with 232 infected patients (cases) and 651 non-infected ones (control). %However, as the exclusion criteria are not very strict, the average fraction of missing data is as high as 86 $\%$. 

\subsection{Strategies for dealing with missing data}
%The recorded EHR data can vary greatly from patient to patient, making it an achievement for temporal analysis. Furthermore, most statistical and machine learning methods work with complete datasets, although several approaches have been proposed to deal with observations at non regular sampling. Following~\cite{bahadori2012}, these methods can be categorized into: (1) smoothing or interpolation techniques; (2) spectral analysis tools such as wavelets or Lomb-Sargle Periodogram; and (3) kernel methods. 

Since EHR data might be missing for several different reasons, 
in many cases the missingness mechanism cannot be described exclusively as either \emph{missing completely at random} (MCAR), \emph{missing at random} (MAR) or \emph{missing not at random} (MNAR), but rather as a  combination of all these three traditional schemes~\cite{ wells2013strategies, hu2017strategies}. Indeed, one reason for missing data is \emph{lack of documentation}, which occurs for example when a clinician orders a blood test but the test for some reason is not performed, or it is performed but not documented because of an error by the physician that performs the test or the data is lost during extraction.
Such type of missing data is usually MCAR or MAR.  However, missingness can also be caused by \emph{lack of collection}. This happens, for example, when the clinician that is treating the patient thinks the health condition of the patients is so good that there is no reason to order a blood test on that particular day. In this case, data are MNAR. 

Most machine learning methods, and in particular  discriminative learning algorithms, work only with complete datasets~\cite{little2014statistical}. However, in a clinical setting, creating a complete dataset by simply discarding patients with missing data may lead to incorrect assessments or prognostics, since the fraction of missing data is typically large~\cite{soguero2016predicting}. Hence, one could end up discarding a lot of information and get a weak model. Moreover, complete case analysis as a result of discarding data only give unbiased predictions if the missingness mechanism is MCAR~\cite{rubin1976inference, schafer1997analysis}.

As an alternative, a preprocessing step involving \emph{imputation} of missing values with some estimated value is common. For example one could fill the missing values for each variable of interest with the mean or median value of the observed samples. 
A simple, but sometimes efficient approach, is to impute all missing values with zeros. Other so-called \emph{single imputation} methods include machine learning based methods such as multilayer perceptrons, self-organizing maps, k-nearest neighbors, recurrent neural networks and regression-based imputation~\cite{garcia2010pattern, RAHMAN2015198}.

The strategies for handling missing data discussed so far are general and can be applied to both vectorial and time series data. On the other hand, methods that only apply to time series data include to impute missing values using smoothing and interpolation techniques such as
the well-known last observation carried forward (LOCF) scheme that imputes the last non-missing value for the following missing values. Further approaches for MTS are linear interpolation,  moving average and Kalman smoothing, to name a few~\cite{ENGELS2003968}. The list of possible imputation methods for MTS data is almost endless and a comprehensive overview of all these is beyond the scope of this paper. For a more detailed overview, we refer the interested reader to~\cite{garcia2010pattern, donders2006review, durbin2012time}.

A drawback common to all imputation methods discussed above is that the information about which values are actually missing is lost. 
Moreover, imputation typically also introduces additional bias into the data due to strong assumptions made by the imputation method. 
For example mean imputation often leads to biased (shorter) estimates of the distances between data points than what they actually are~\cite{hu2017strategies}.
To resolve the bias problem it has been suggested to correct for the bias by introducing binary indicator variables to account for the missingness pattern~\cite{hu2017strategies}.  An additional problem with single imputation is that the uncertainty associated with the missing values is ignored since they are replaced with ``certain" estimates, which in turn may lead to smaller estimated standard errors than the true standard errors. 
\emph{Multiple imputation}~\cite{white2011multiple} resolves this problem by estimating the missing values multiple times and thereby creating multiple complete datasets. Thereafter for example a classifier is applied to all datasets and the results are combined to obtain the final predictions. However, applying multiple imputation to MTS in a clustering setting is a non-trivial task that involves several challenges~\cite{basagana2013framework}. %Multiple imputation provides unbiased results if the imputation model is correct and the missingness mechanism is MAR~\cite{schafer1997analysis}.

%The issues introduced by doing imputation can affect the analysis considerably.
Despite that multiple imputation and other imputation methods can give satisfying results in some scenarios,  these are ad-hoc solutions that lead to a multi-step procedure where the missing data are handled separately and independently from the rest of the analysis~\cite{wells2013strategies}. This is not an optimal solution, and therefore several research efforts have been devoted over the last years to process incomplete temporal data without relying on imputation~\cite{DBLP:journals/corr/ChePCSL16, bianchi2018time, DBLP:journals/corr/abs-1711-06516, mikalsen2016learning, pmlr-v56-Lipton16, Marlin:2012:UPD:2110363.2110408}.
In this regard, powerful kernel methods have been recently proposed, of which the  TCK and LPS are prominent examples. 
Even though there are many similarities between these two kernels, the way missing data are dealt with is very different. In LPS  the missing data handling abilities of decision trees are exploited. Along with ensemble methods, fuzzy approaches and support vector solutions, decision trees can be categorized as \emph{machine learning approaches for handling missing data}~\cite{garcia2010pattern}. Common to these approaches is that the missing data are handled naturally by the machine learning algorithm. One can also argue that the way missing data are dealt with in the TCK belongs to this category, since an ensemble approach is exploited. However, it can also be categorized as a \emph{likelihood-based approach} since the underlying models in the ensemble are Gaussian mixture models. In the likelihood-based approaches the full, incomplete dataset is analyzed using maximum likelihood (or maximum a posteriori, equivalently), typically in combination with the expectation-maximization (EM) algorithm~\cite{schafer2002missing, little2014statistical}. 

The main advantage of these methods, compared to imputation methods, is that the missing data are handled automatically and no additional tasks are left to the user. For example in multiple imputation, a careful selection of the imputation model and other variables is needed to do the imputation~\cite{schafer2002missing},  which in particular in an unsupervised setting can turn out to be problematic. Moreover, similarly to multiple imputation, unbiased predictions are guaranteed if data are MAR.

A more detailed description of the TCK and LPS kernels is provided in the next subsection, along with a description of the other kernels for MTS used in this work.

%##########################
% LPS and TCK
%###########################################################
\subsection{Multivariate time series kernels}
%\commentR{Make it clearer in this section why kernel \emph{matrices} enter the picture?} 
Kernel methods have been of great importance in machine learning for several decades and have applications in many different fields~\cite{Jenssen2010,Jenssen2013,camps2009kernel,soguero2016support}.
Within the context of time series, a \textit{kernel} is a similarity measure that also is positive semi-definite~\cite{shawe2004kernel}. 
Once defined, such similarities between pairs of time series may be utilized in a wide range of applications, such as classification or clustering, benefiting from the vast body of work in the field of kernel methods.

%%%%%%%%%%%%%%%%%%%%%%%%%%%%%%%%%%%%%%%
% linear kernel
%%%%%%%%%%%%%%%%%%%%%%%%%%%%%%%%%%%%%%%%%%%%%%%%%%%%%%5

\paragraph{Linear kernel}
%\subsubsection{Linear kernel}
The simplest of all kernel functions is the linear kernel, which for two data points represented as vectors, $x$ and $y$, is given by the inner product $\langle x, y \rangle$, possibly plus a constant $c$. 
One can also apply a linear kernel to pairs of MTS once  they are unfolded into vectors. However, by doing so the information that they are MTS and there might be inherent dependencies in time and between  attributes, is then lost. 
Nevertheless, in some cases such a kernel can be efficient, especially if the MTS are short~\cite{chen2013model}.
If the MTS contain missing data, the linear kernel requires a preprocessing step involving e.g. imputation.

%%%%%%%%%%%%%%%%%%%%%%%%%%%%%555
%GAK 
%%%%%%%%%%%%%%%%%%%%%%%%%%%%%%%%%55

\paragraph{Global alignment kernel}
%\subsubsection{Global alignment kernel}
The most widely used time series similarity measure is \emph{dynamic time warping} (DTW) \cite{Berndt:1994:UDT:3000850.3000887}, where the similarity is quantified as the alignment cost between the MTS. More specifically, in DTW the time dimension of one or both of the time series is warped to achieve a better alignment. 
Despite the success of DTW in many applications, similar to many other similarity measures, it is non-metric and therefore cannot non-trivially be used to design a positive semi-definite kernel~\cite{ marteau2015recursive}. Hence, it is not suited for kernel methods in its original formulation.
Because of its popularity there have been attempts to design kernels exploiting the DTW. For example Cuturi et al. designed a DTW-based kernel using global alignments~\cite{cuturi2007kernel}. 
An efficient version of the global alignment kernel (GAK) is provided in~\cite{cuturi2011fast}. 
The latter has two hyperparameters, namely the kernel bandwidth and the triangular parameter. 
These are usually set using some heuristics.
GAK does not naturally deal with missing data and incomplete datasets, and therefore also requires a preprocessing step involving imputation. 

%###################################################
%TCK
%###############################################################

\paragraph{Time series cluster kernel}
%\subsubsection{Time series cluster kernel}
\label{sec:TCK}

The TCK is based on an ensemble learning approach~\cite{Strehl:2003} wherein the robustness to hyperparameters is ensured by joining the clustering results of many Gaussian mixture models (GMM) to form the final kernel. Hence, no critical hyperparameters have to be tuned by the user, and the TCK can be learned in an unsupervised manner. 
To ensure robustness to sparsely sampled data, the GMMs that are the base models in the ensemble,  are extended using informative prior distributions such that the missing data is explicitly dealt with.

More specifically, the TCK matrix is built by fitting GMMs to the set of MTS for a range of number of mixture components. The idea is that by generating partitions at different resolutions, one can capture both the local and global structure of the data. 
Moreover, to capture diversity in the data, randomness is injected by for each resolution (number of components) estimating the mixture parameters for a range of random initializations and randomly chosen hyperparameters. In addition, each GMM sees a random subset of attributes and segments in the MTS. 
The posterior distributions for each mixture component are then used to build the TCK matrix by taking the inner product between all pairs of posterior distributions.
Finally, given an ensemble of GMMs, the TCK is created in an additive way by using the fact that the sum of kernels is also a kernel. In this work, we have modified the kernel slightly from the way it was originally proposed in~\cite{mikalsen2017time} by normalizing the vectors of posteriors to have unit length in the $l_2$-norm. This provides an additional regularization that may increase the generalization capability of the learned model. A more detailed description of the method is provided in \ref{appendix: TCK}. 

%########################################################################################
%LPS
%####################################################################################
\paragraph{Learned pattern similarity}
%\subsubsection{Learned pattern similarity}

LPS is a similarity measure that satisfies the requirements of a kernel, as shown in \cite{mikalsen2017time}, which can naturally deal with MTS. 
Similar to the TCK, the LPS is also based on extracting random segments. 
Additionally, the LPS is similar to the TCK in the sense that one in an unsupervised way can learn a similarity between time series that is robust to hyperparameter choices and can deal with missing data using the missing data handling properties of tree-based learning. It generalizes the well-known autoregressive models~\cite{shumway} to local autopatterns using multiple lag values for autocorrelation. These autopatterns are supposed to capture the local dependency structure in the time series and are learned using a tree-based learning strategy. 

More specifically, a time series is represented as a matrix of segments. Randomness is injected to the learning process by randomly choosing time segment (column in the matrix) and lag $p$ for each tree in the random forest. A bag-of-words type compressed representation is created from the output of the leaf-nodes for each tree. The final time series representation is created by concatenating the representation obtained from the individual trees, which in turn are used to compute the similarity using a histogram intersection kernel~\cite{barla2003histogram}.

Given two MTS $X^{(n)}$ and $X^{(m)}$, a formal expression for the LPS-kernel is 
\begin{equation} \label{eq: LPS}
    K(X^{(n)},X^{(m)}) = \frac{1}{R J} \sum\limits_{k=1}^{R J} \min (h^n_k, h^m_k),
\end{equation}
where $h^n_k$ is the $k$th entry of the concatenated bag-of-words representation $H(X^{(n)})$.  More precisely, $H(X^{(n)})$ is a concatenation of $R$-dimensional frequency vectors of instances in the terminal nodes from all $J$ trees.

 %Moreover, since both methods are valid kernels and can therefore be applied for classification, clustering, and dimensionality reduction within the theoretically well-understood framework of kernel methods. %However, the experiments in \cite{mikalsen2017time} indicated that for shorter time series with much missing data the TCK is superior to the LPS. 
%More detailed descriptions of the two methods are provided in \ref{appendix: TCK} and \ref{appendix: LPS}, respectively. \textbf{I guess the appendices are not needed}

%\subsection{Spectral clustering}
%In \emph{clustering} one tries to group a set of objects such that objects in the same clusters (groups) are more similar to each other than the objects in other clusters. \emph{Spectral clustering} is a family of clustering methods that utilizes the spectrum of some similarity matrix of the data~\cite{ng2002spectral}. Most commonly in this clustering method one employs a two step procedure, where in the first step a lower dimensional representation is computed based on the eigenvalues and eigenvectors of the similarity matrix and thereafter k-means is applied to the new representation. A detailed explanation of such a two-step procedure is given in \cite{ng2002spectral}. In this work we will employ the similar scheme, but with the exception that we use kernel principal component analysis (kPCA)~\cite{scholkopf1997kernel} to compute the low dimensional representation and use the TCK to compute the kernel matrix.

%\subsection{Time series clustering using spectral clustering}
\subsection{Model development}
The kernels that we described in the previous section are used to compute a kernel matrix on a training set, which is created by randomly selecting 80 percent of the dataset. The remaining 20 percent is used as test set. 
The LPS and TCK kernels are computed on the incomplete dataset containing missing data, whereas the GAK and linear kernel cannot work on incomplete datasets, and we therefore compute these on 6 different complete datasets obtained using mean imputation, LOCF imputation, 0-imputation, and replicates of these corrected for bias. In the bias corrected (BC) datasets we double the number of attributes in each MTS by stacking a binary MTS, representing imputed elements, to the imputed MTS. 
When using imputation of the mean, we calculate the mean for each attribute in the MTS, across all time intervals in the postoperative window and all patients in the training set. If an element is missing in the first time interval, we replace it with the mean when we do LOCF imputation.

After having computed the different kernels, we take an unsupervised approach to classifying the patients with SSI using the \emph{spectrum} of the kernel matrices. We employ a variant of spectral clustering consisting of two steps, namely kPCA followed by k-means. In the first step, kPCA with the learned MTS kernel is used to compute a low dimensional representation of the MTS.
Thereafter we cluster the learned representations using \emph{k-means}. We assume that the number of clusters is known and set it to 2.
Out-of-sample data are assigned to clusters according to the cluster labels of the \emph{k-nearest neighbors} (kNN) in the training set. 
The processing pipeline we have described here is also illustrated in Fig.~\ref{fig:Scheme}.

\subsection{Model  evaluation}
The different models are evaluated both on the training and test set. Because of the imbalanced classes we decide to use \emph{F1-score}~\cite{hripcsak2005agreement} instead of accuracy as performance measure. 
F1-score is a function of two metrics, namely \emph{precision} and \emph{recall}. 
These two metrics are also commonly referred to as positive predictive value and sensitivity, respectively. Precision is the fraction of true positives (have infection) among all those that are classified (clustered) as positive cases, whereas recall is the fraction of positive cases in the gold standard classified as positive. F1-score can be expressed in terms of these two metrics as follows:
\begin{equation}
    F_1 = \frac{\text{recall} \times \text{precision}}{\text{recall} + \text{precision}}
\end{equation}
In order to adapt this to an unsupervised regime, we define \emph{clustering F1-score} similarly to how \emph{clustering accuracy} is defined. i.e. we use the permutation of the labels provided by the clustering algorithm that gives the highest score. 
In the following, we refer to both clustering F1-score and classification F1-score simply as F1-score.

The procedure described in the previous section is repeated 10 times such that we can compute both mean and standard errors for the F1-score, i.e. we randomly select 10 different training and test sets (80 and 20 percent, respectively), and repeat the same process on all of them.
In addition, in order to study how stable and robust the different methods are to varying length of the MTS, we vary the size of the postoperative window from 7 to 20 days.

%#####################################
\section{Results} \label{Sec:Results}
%#####################################

The TCK and LPS are run using default hyperparameters~\cite{MikalsenTCK, baydogan2016time}, with the exception for the LPS that we increase the minimal segment length from $5 \%$ to $15 \%$ percent of the length of the MTS to account for the short time series. In accordance with~\cite{Cuturi}, for GAK we set the bandwidth $\sigma$ to two times the median  distance of all MTS  in the training set scaled by the square root of the median length of all MTS, and the triangular parameter~\cite{Cuturi}  to 0.2 times the median length of all MTS. Distances are measured using the canonical metric induced by the Frobenius norm. In the linear kernel we set the constant $c$ to 0.

\begin{figure}[!t]
    \centering
    \subfigure%[]
    {
        \includegraphics[width=0.48\linewidth]{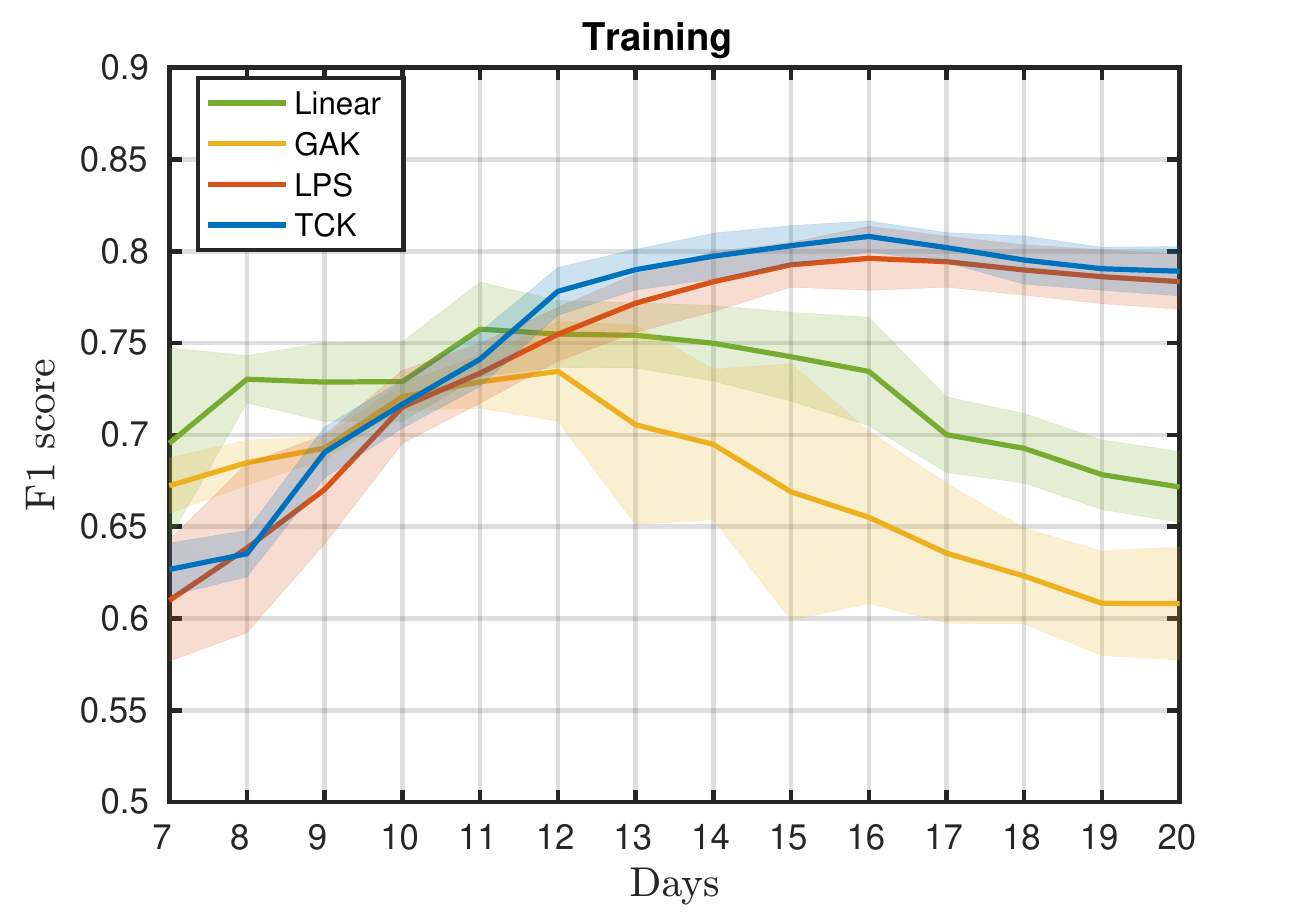}
        \label{fig: train clust}
    }%
    \subfigure%[]
    {
        \includegraphics[width=0.48\linewidth]{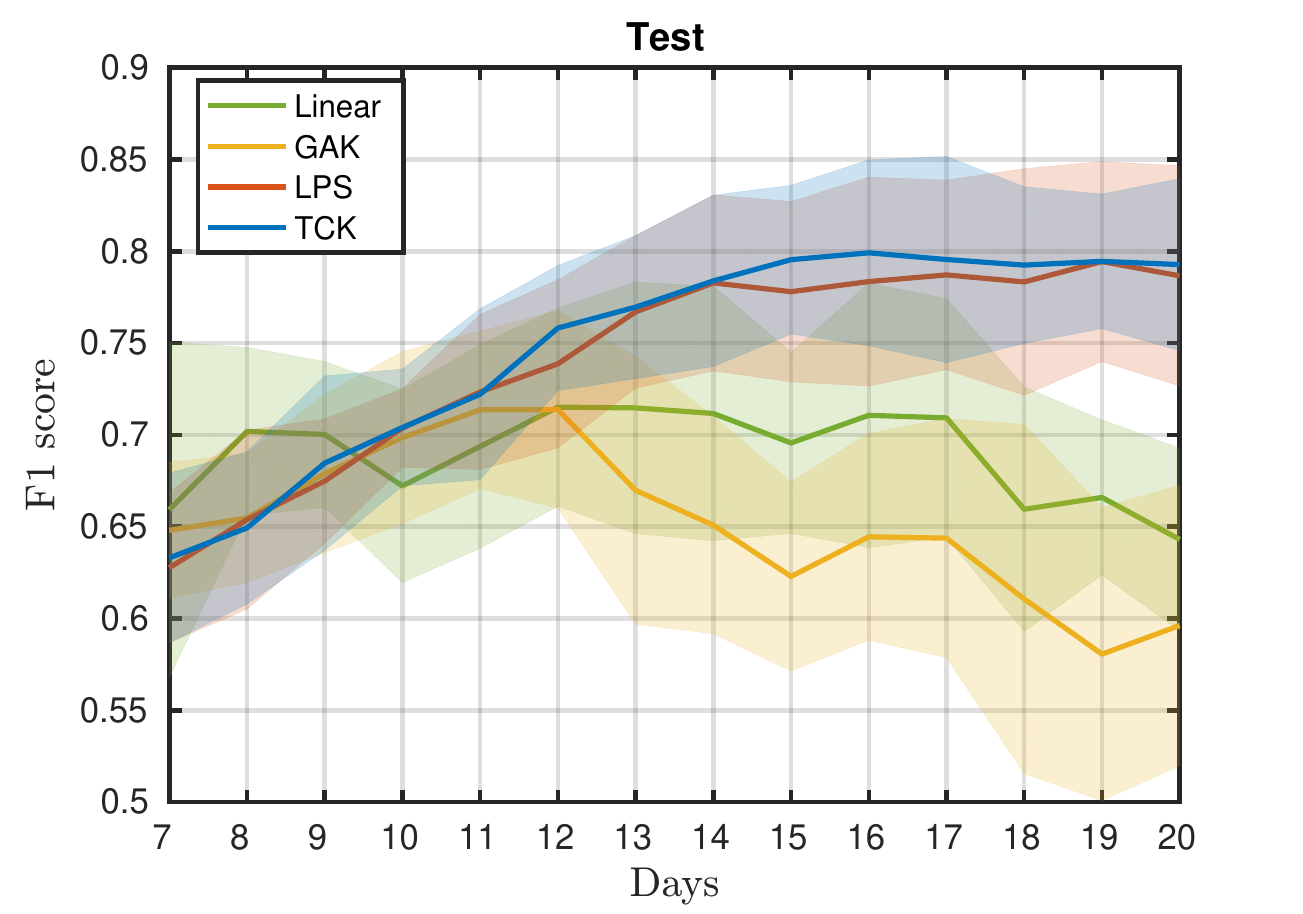}
        \label{fig: test clust}
    }%    
    \caption{Mean F1-score over 10 runs on training (left) and test set (right) and standard errors obtained using four different MTS kernels, followed by kPCA and k-means. Out-of-sample data is clustered using a kNN classifier. The green line represents a linear kernel, yellow the global alignment kernel, red the learned pattern similarity kernel and blue the time series cluster kernel.  }
    \label{fig: clustering results}
\end{figure}

\begin{figure}[!t]
    \centering
    \subfigure%[]
    {
        \includegraphics[width=0.48\linewidth]{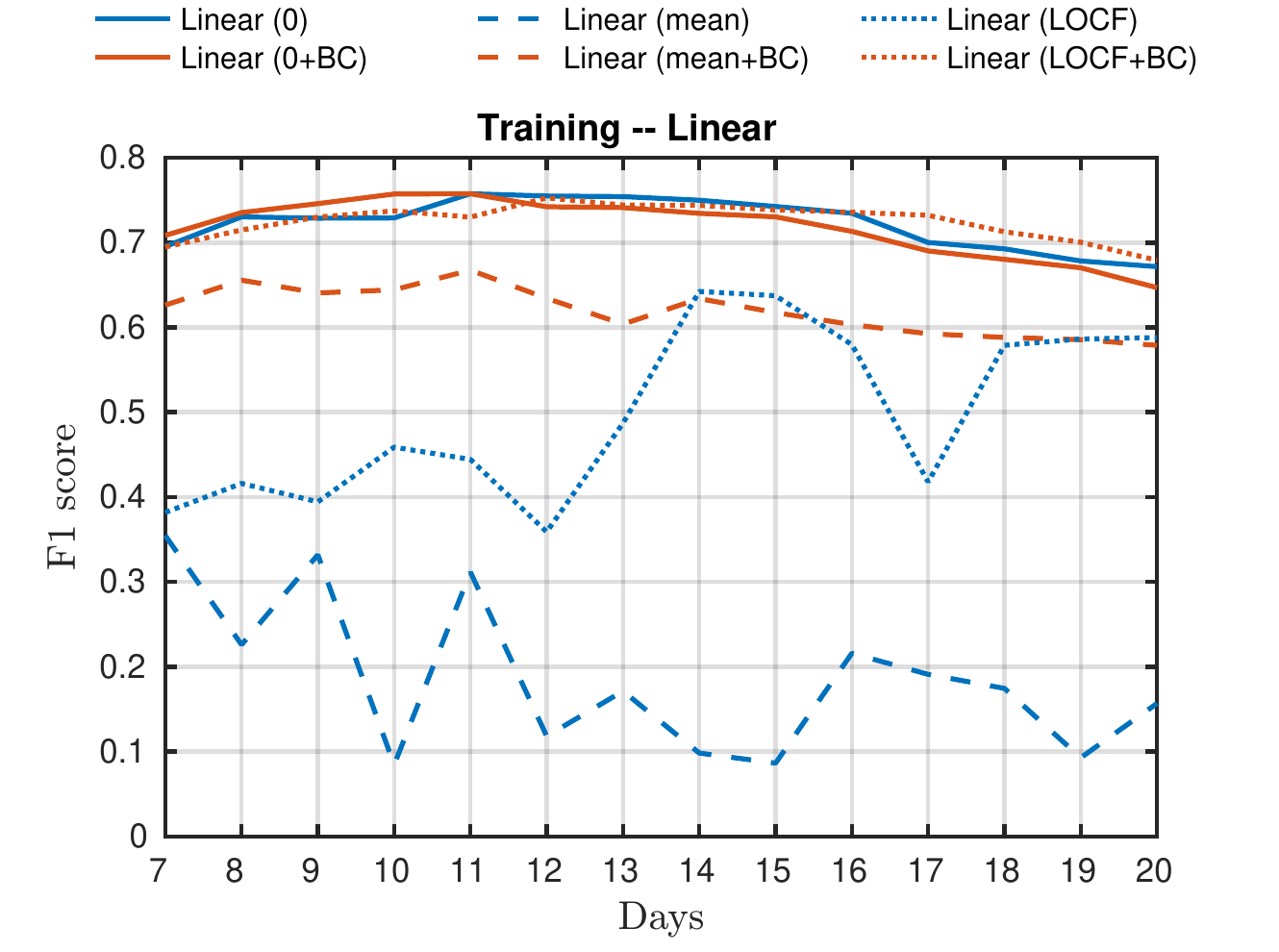}
        \label{fig: train clust lin}
    }%
    \subfigure%[]
    {
        \includegraphics[width=0.48\linewidth]{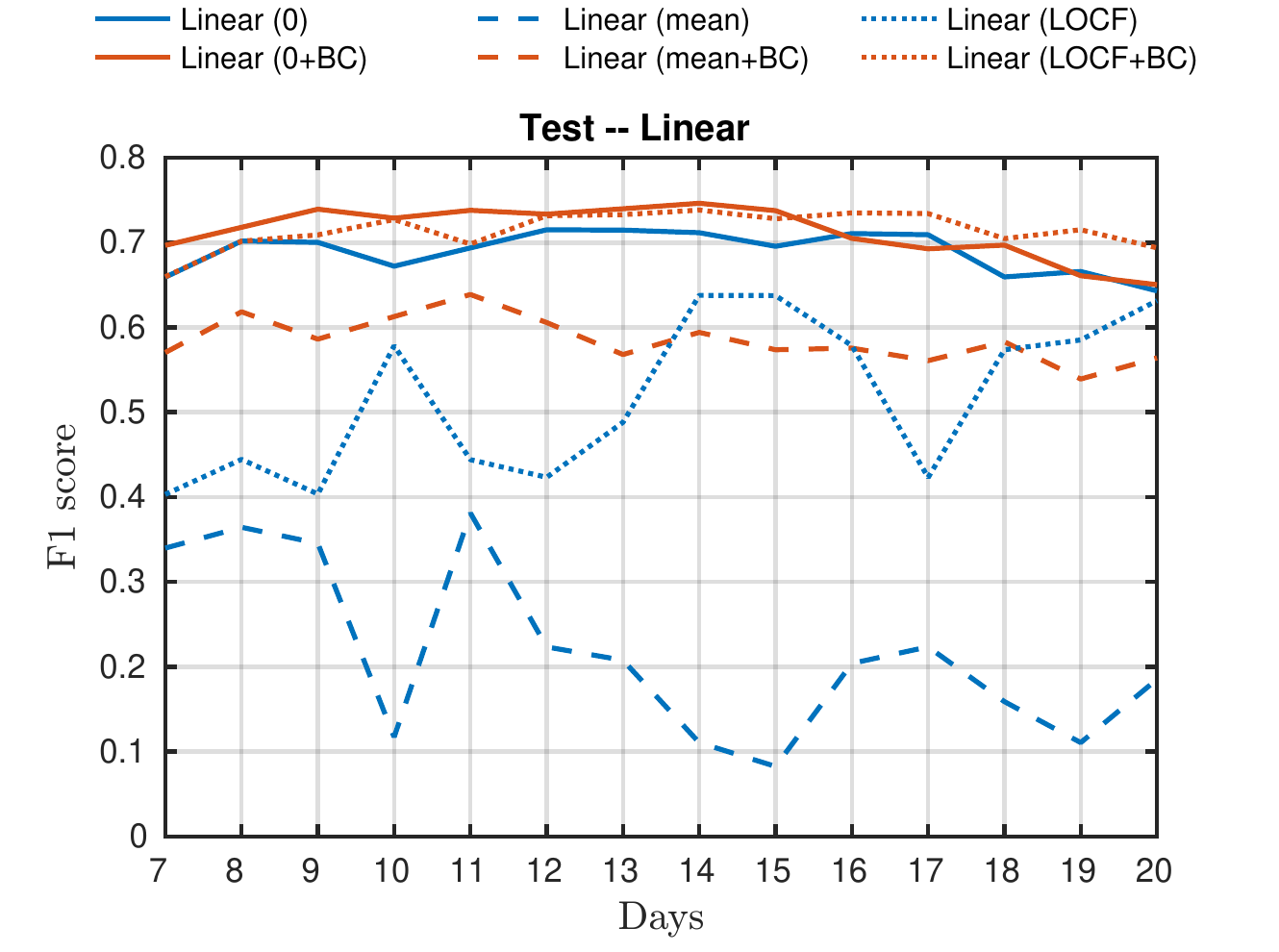}
        \label{fig: test clust lin}
    }%    
    \\
        \subfigure%[]
    {
        \includegraphics[width=0.48\linewidth]{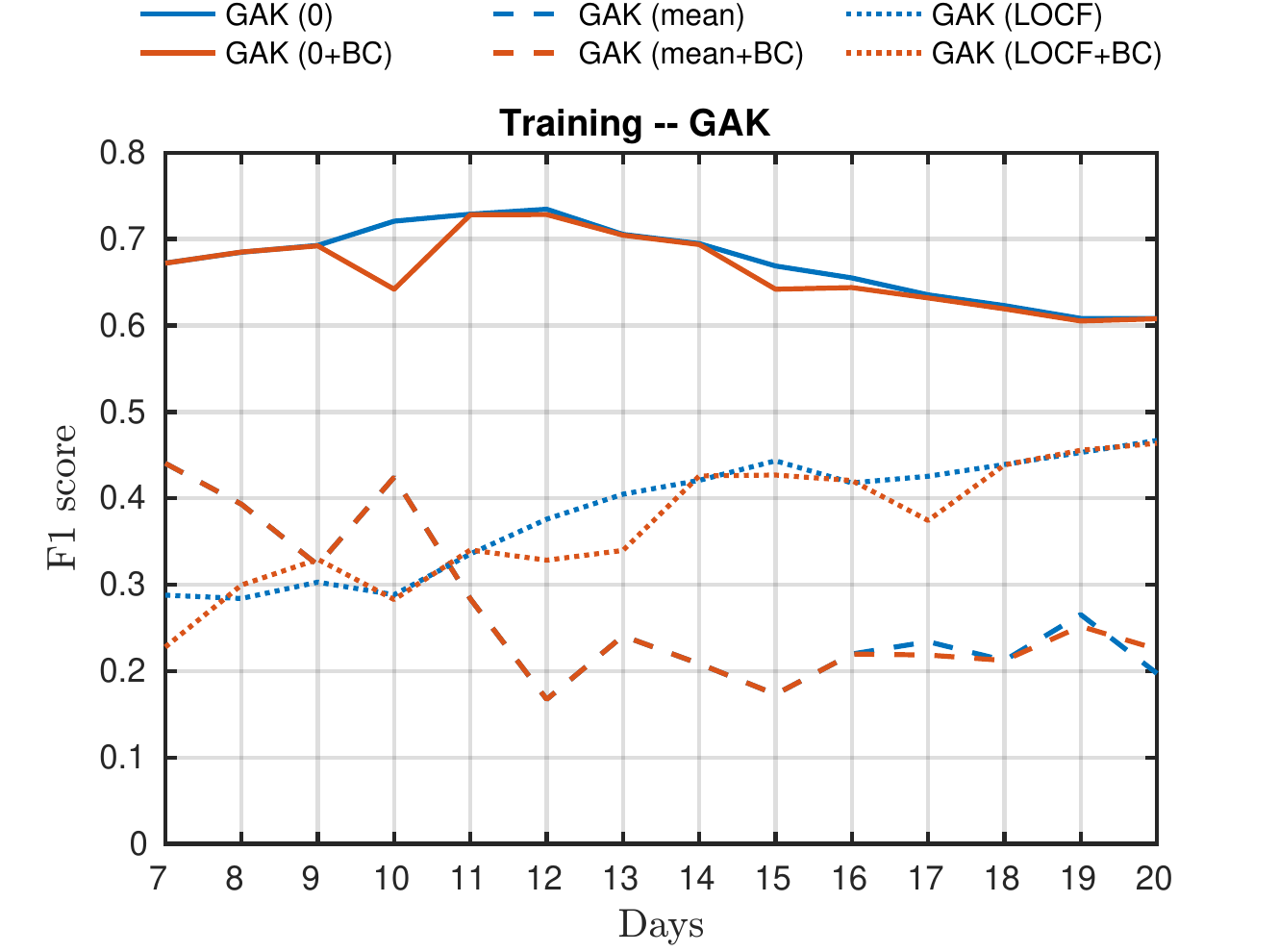}
        \label{fig: train clust gak}
    }%
    \subfigure%[]
    {
        \includegraphics[width=0.48\linewidth]{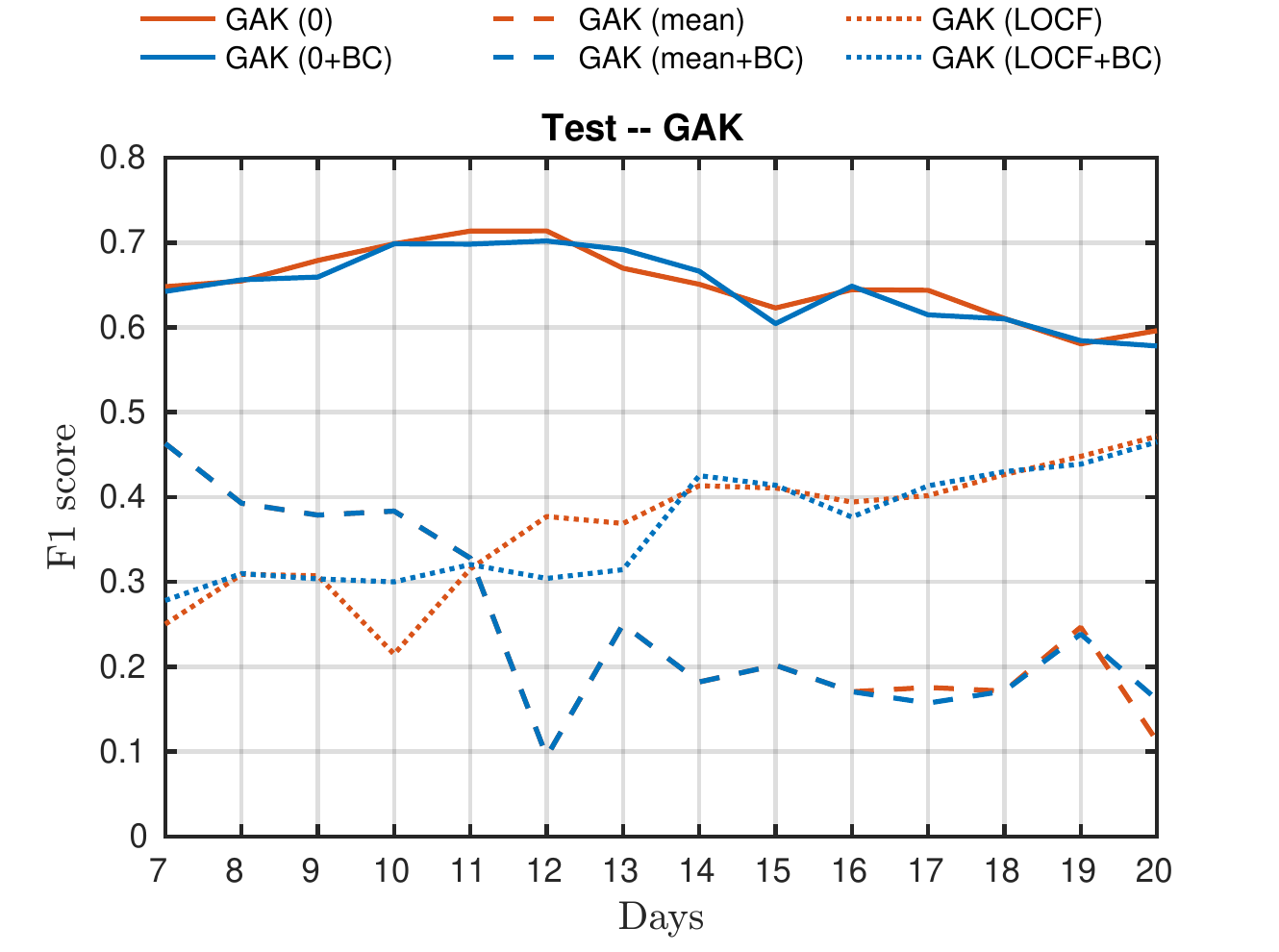}
        \label{fig: test clust  gak}
    }%  
    \caption{Mean F1-score over 10 runs on training (left) and test set (right) obtained using the linear kernel and GAK  with six different imputation methods, followed by kPCA and k-means. Out-of-sample data is clustered using a kNN classifier. Standard errors are not shown for increased readability.}
    \label{fig: clustering results lin vs gak}
\end{figure}

Fig.~\ref{fig: clustering results} shows mean F1-score  over 10 runs on the training (left) and test set (right), and standard errors, obtained using LPS and TCK kernels, followed by kPCA to 10 dimensions and k-means, where test data are clustered using a kNN classifier with $k=5$ and the cluster assignments as labels. %Initial experiments showed that the clustering results are stable to varying size of these parameters.
Initial experiments showed that the clustering results are stable to varying values of these hyperparameters.
For easier comparison, in the same figure we have  also added results obtained with the GAK and linear kernel on the imputed dataset that gives the highest F1-score,
namely 0-imputation, whereas the results for all 6 complete datasets are shown in Fig.~\ref{fig: clustering results lin vs gak}.  

We observe that the two kernels, TCK and LPS, that can explicitly deal with the missing data, give very similar results both on the training and test set. 
When the postoperative window is 7 days the two methods yield an F1-score of approximately 0.63, then the performance increases almost linearly from day 7 to 15 where it stabilizes around a F1-score of 0.80. Further, it can be seen that TCK, LPS and GAK perform worse than the linear kernel when the postoperative window is short ($< 10$ days).
However, as the size of the postoperative window increases, the relative performance of the TCK and LPS with respect to both GAK and the linear kernel improve. Even if the differences between LPS and TCK are small, it can be seen that the latter yields slightly better performance (in average) for all sizes of the postoperative window both in training and test. Moreover, the standard errors with the TCK are smaller, in particular in training when the postoperative window is short.
This is particularly important in unsupervised frameworks like the one we are proposing.

For the two kernels that work on the imputed data,  we observe that
GAK (0 and 0+BC) and Linear (0, 0+BC, LOCF+BC) perform quite similarly, and these are the five imputed data methods that give the best performance.  
A common pattern for these five methods, and especially for GAK (0 and 0+BC), is that the F1-score increases when the postoperative window is increased from 7 to around 11 days (at least on the training set), but then the F1-score slowly starts decrease after that. The performance of the seven other imputation methods is considerably worse.

\begin{figure}[!t]
    \centering
    \subfigure[TCK]
    {
        \includegraphics[width=0.48\linewidth]{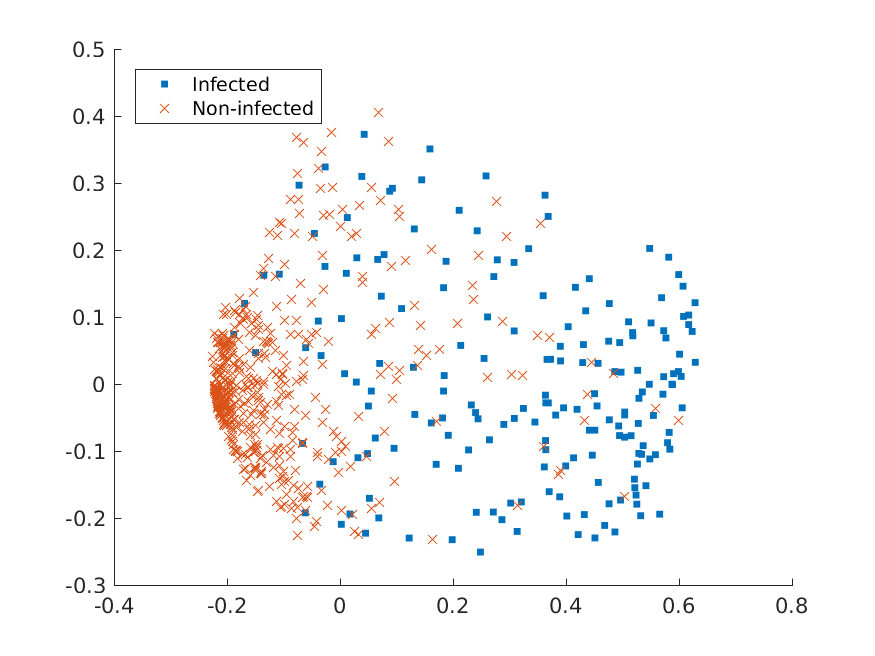}
        \label{fig: KPCA TCK 20}
    }%
    \subfigure[LPS]
    {
        \includegraphics[width=0.48\linewidth]{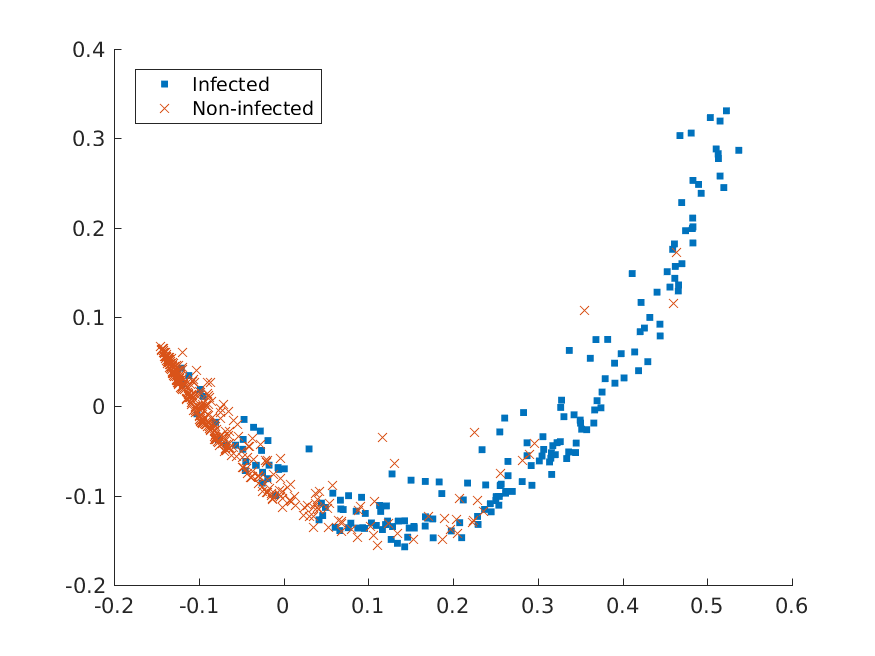}
        \label{fig: KPCA LPS 20}
    }  
    \subfigure[GAK]
    {
        \includegraphics[width=0.48\linewidth]{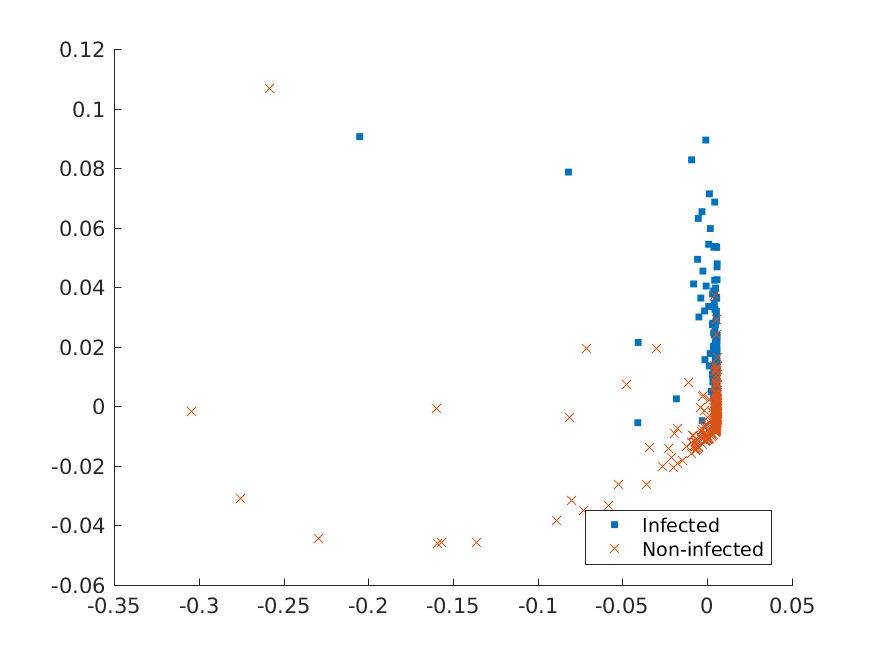}
        \label{fig: KPCA GAK 20}
    }%
    \subfigure[Linear kernel]
    {
        \includegraphics[width=0.48\linewidth]{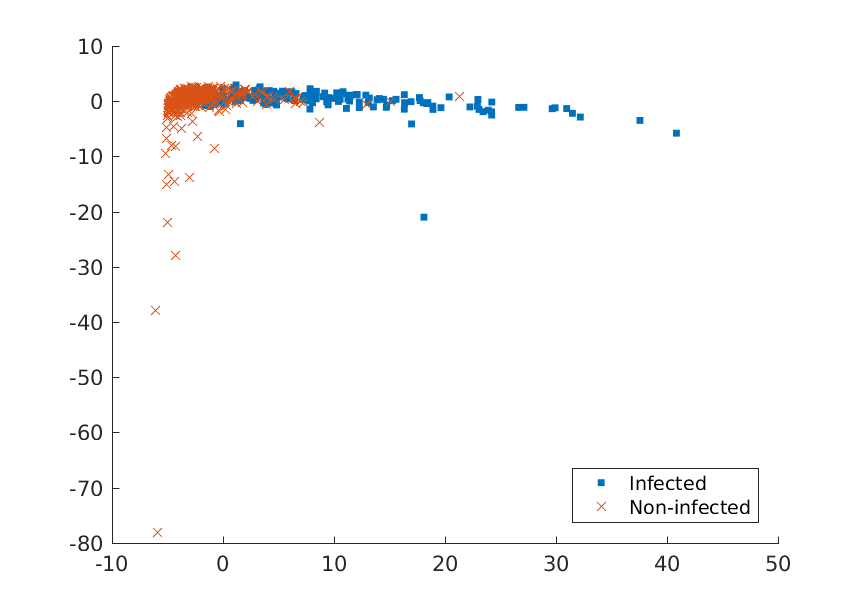}
        \label{fig: KPCA linear 20}
    }%       
    \caption{Plot the kPCA representation corresponding to the two largest eigenvalues obtained using four different kernels on the training set with postoperative window size equal to 20. The patients are color coded according to their labels (infected with SSI or not). }
    \label{fig: KPCA}
\end{figure}

To further investigate the differences between LPS and TCK, beside F1-score, in Fig.~\ref{fig: KPCA TCK 20} and Fig.~\ref{fig: KPCA LPS 20} we show the kPCA representation corresponding to the two largest eigenvalues obtained using these two kernels on the training set with postoperative window size equal to 20. Interestingly the representations created by the LPS and TCK are very different.
The LPS  has a clearer manifold structure, whereas the TCK  is more spread out in the plane. Even though it is difficult to argue that one of these two representations is superior to the other, the TCK at least more clearly reflects that the cohort of patients is very diverse and complex because of large individual differences. 
To better understand the performance of the GAK and linear kernel we also plot the 2D kPCA representation obtained on 0-imputed data in Fig.~\ref{fig: KPCA GAK 20} and Fig.~\ref{fig: KPCA linear 20}. We note that these are heavily influenced by outliers. %\commentR{Possible to analyse the issue of outliers more? Are these outliers because of heavy imputation?}. 
Apart from the outliers, the other datapoints become very compact and close to each other, and it is therefore not strange that the clustering algorithm does not identify the groups correctly.

%We also note that, in general, the performance on the test set is very similar to the performance on the training set for all kernels and all methods when dealing with missing data.

\begin{figure}
    \centering
    \subfigure%[]
    {
        \includegraphics[width=0.48\linewidth]{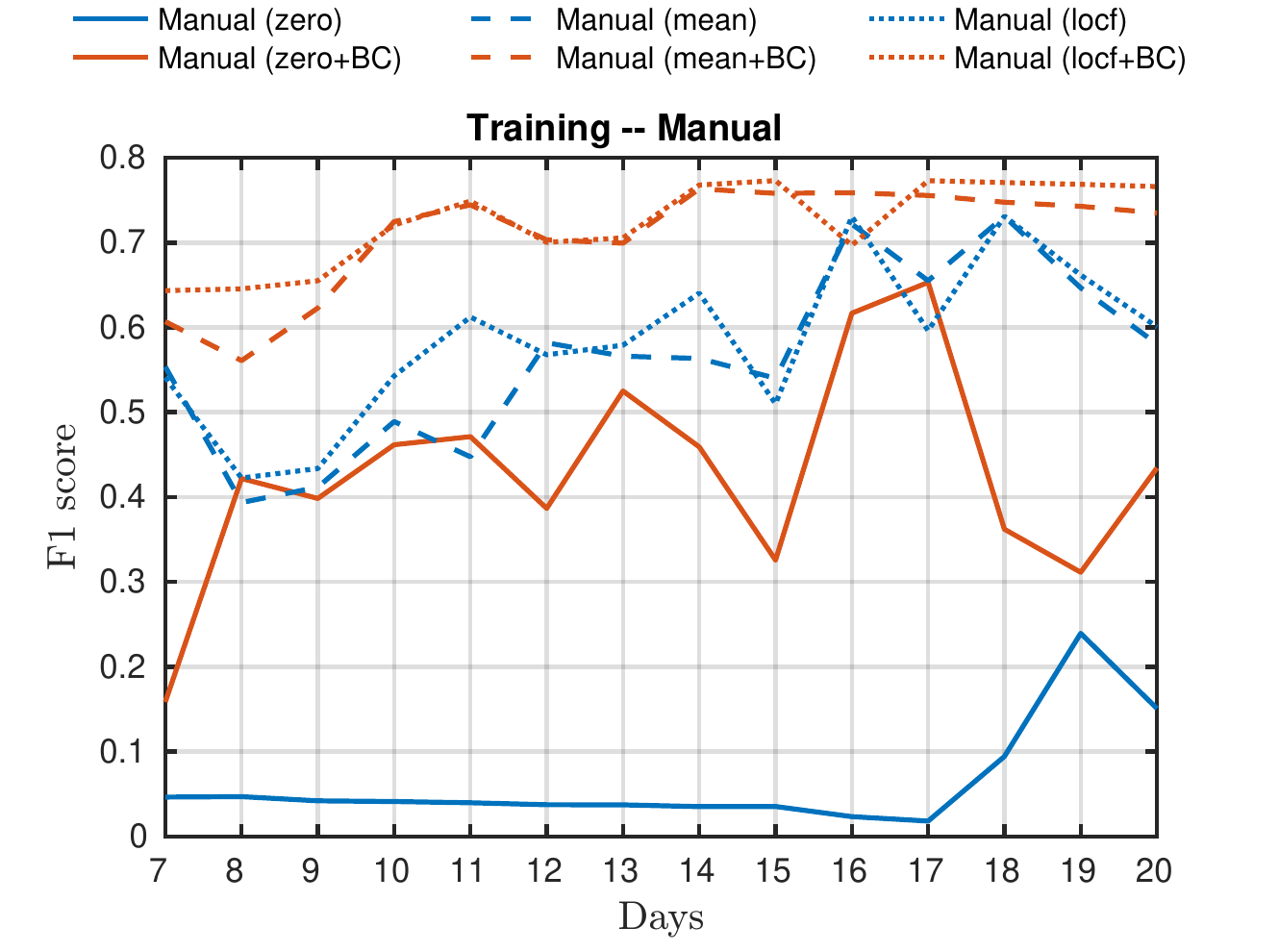}
        \label{fig: manual tr}
    }%
    \subfigure%[]
    {
        \includegraphics[width=0.48\linewidth]{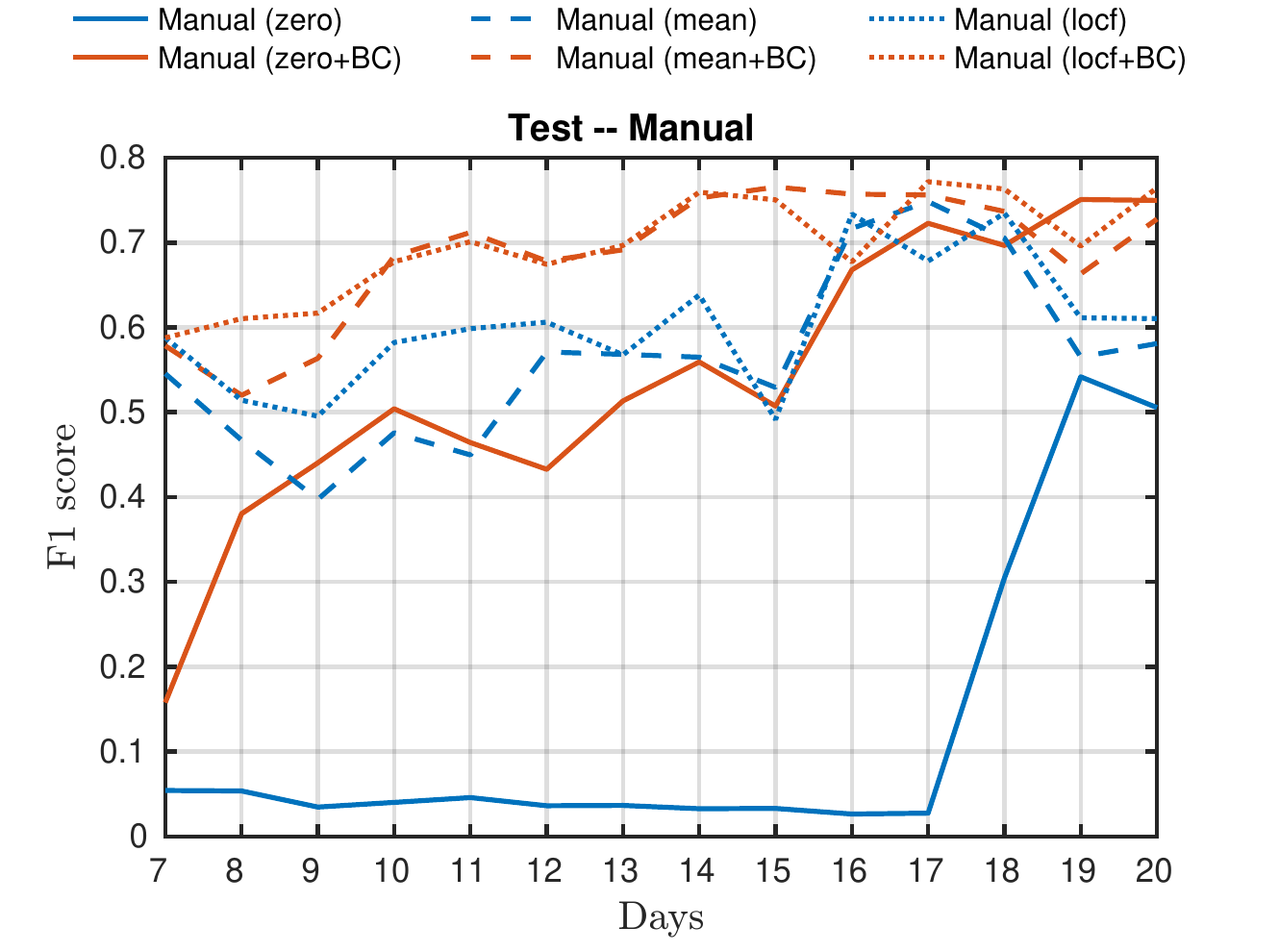}
        \label{fig: manual te}
    }%    
    \caption{Mean F1-score over 10 runs on training (left) and test set (right) obtained using the static, manually extracted features in combination with a linear kernel and six different imputation methods, followed by kPCA and k-means. Out-of-sample data is clustered using a kNN classifier. Standard errors are not shown for increased readability. }
    \label{fig: clustering results manual}
\end{figure}

As a baseline to compare the performance of our method, we follow the idea proposed by~\cite{hu2017strategies} and  compute higher level non-temporal features from the longitudinal data. 
Specifically, in addition to the average value of each of the blood tests, we compute two extreme values (the maximum and minimum) over the postoperative window. In this setting we consider a missing value as a specific blood test that is missing entirely during the postoperative window. This baseline cannot naturally deal with missing data and therefore we use the same six missing data imputation strategies as described for the temporal features. We apply a linear kernel to the manual features and then we follow the same scheme as in the temporal case.
Fig.~\ref{fig: clustering results manual} shows mean F1-score over 10 runs on training (left) and test set (right) and standard errors obtained using this baseline on the six imputed datasets.
Similarly to the results obtained with the linear kernel and GAK on the temporal data, the results obtained with this baseline heavily depend on the type of imputation method, and are also very fluctuating as function of length of the postoperative window.  %\textbf{not robust bla bla, highest F1-score slightly above 0.75}

\begin{figure}
    \centering
    \subfigure%[]
    {
        \includegraphics[width=0.48\linewidth]{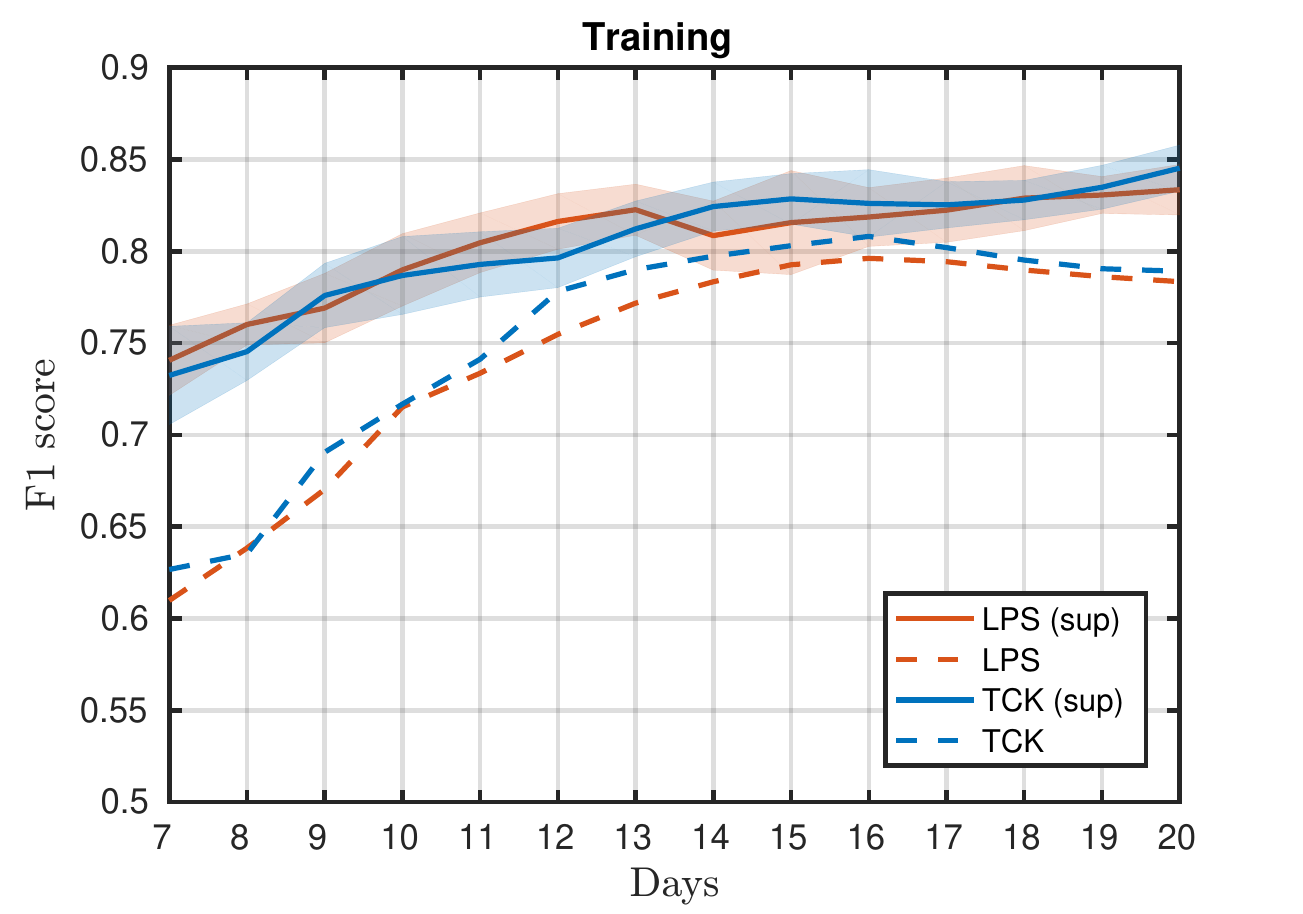}
        \label{fig: train class}
    }%
    \subfigure%[]
    {
        \includegraphics[width=0.48\linewidth]{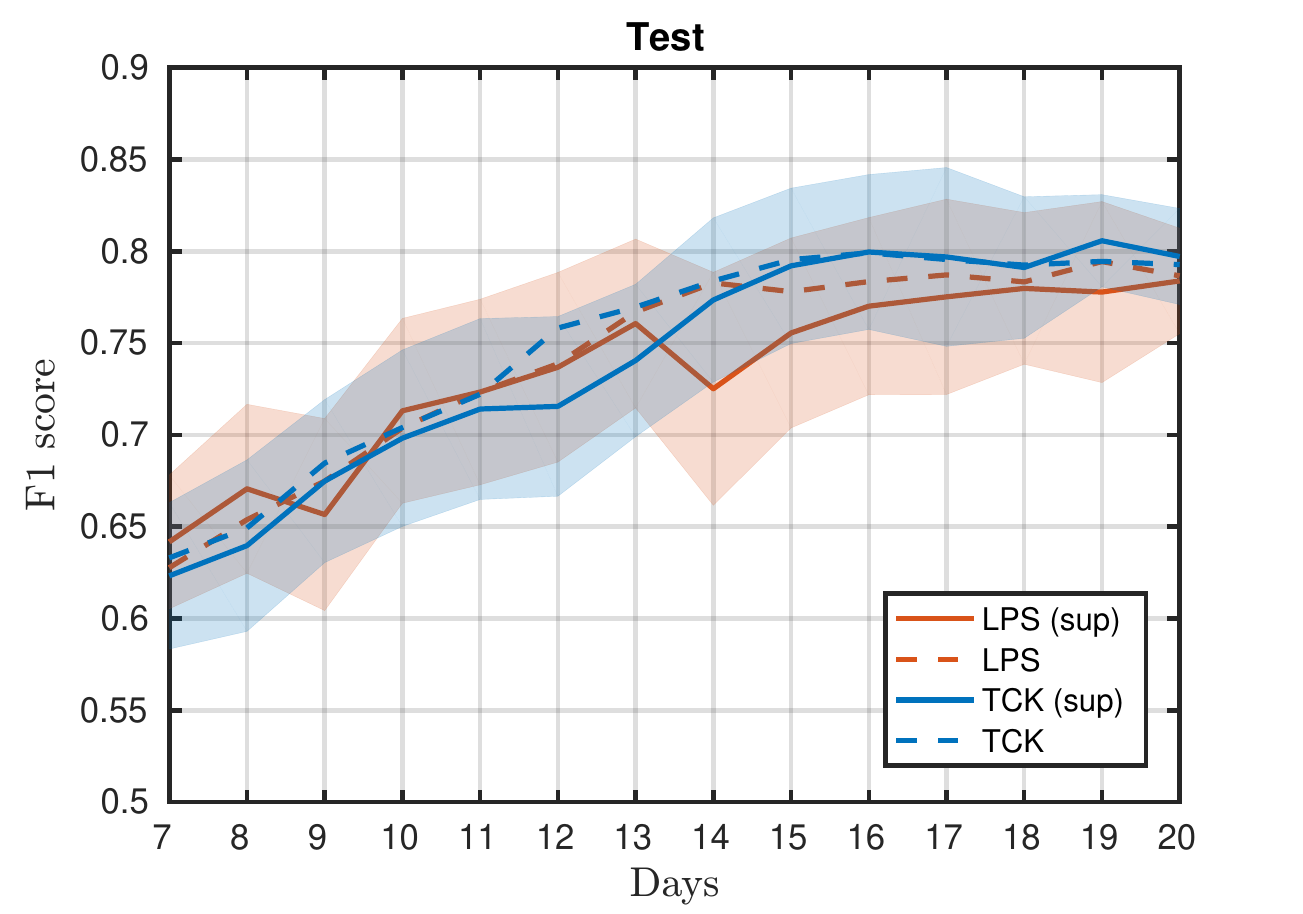}
        \label{fig: test class}
    }%    
    \caption{Mean F1-score over 10 runs on training (left) and test set (right) and standard errors obtained using the TCK and LPS kernels, followed by kPCA to 10 dimensions and a kNN-classifier with  with $k=5$. The dashed lines represent the mean clustering F1-score (from Figure~\ref{fig: clustering results}).}
    \label{fig: classification results}
\end{figure}

In addition to comparing different kernels and methods for dealing with missing data, to see how well the unsupervised classification part works, we also benchmark the proposed unsupervised framework against a supervised baseline where the two first steps are the same, namely to compute the kernel and thereafter do kPCA. However, on the 10 dimensional kPCA representation we employ a kNN classifier with $k=5$ using the true labels. 
Figure~\ref{fig: classification results} shows mean classification F1-score and standard errors over the same 10 randomly drawn training (left) and test sets (right). 
Also with this baseline the performance of the TCK and LPS kernel is very similar. Moreover, in general, the supervised baseline performs better on the training set than the unsupervised method. This is expected. However, on the test set, the F1-scores obtained using the supervised baseline is almost identical to the proposed method for all sizes of the postoperative window. This implies that labelling is not a required task for identifying patients with surgical site infection from blood samples.

\section{Discussion}
\label{sec: discussion}
%\commentR{Out of sample any point in discussing for LPS vs TCK?} %\textbf{not sure if I understand what you mean} \\ \commentR{If we take new measurements, do we need to recompute everything? Any difference between the models? Not a big issue.}

In this work, we have proposed a framework for SSI identification based on secondary use of EHR data.
Our first objective was to alleviate the problem of getting access to labelled EHR data.
The results presented in the previous section clearly show that the proposed framework can detect accurately SSI, without relying on supervised training. In fact, the supervised baseline did not improve the performance compared to the proposed method.
The second objective has been to deal with missing data effectively.
We tackled this problem by introducing robust time series kernels as a main component in our framework and in the following we discuss our findings.

For the two kernels that work on the imputed data, GAK and linear, the choice of imputation method heavily affects the performance more than the choice of kernel itself. 
Both the linear kernel and GAK give very bad results especially with mean imputation, but also with LOCF imputation. 
In those cases, the mean F1-score is far from smooth over time. 
On the other hand, 0-imputation gives good results compared to the other imputation methods for both kernels.
This can be surprising, since 0-imputation introduces a strong bias because blood test values are positive and therefore the value 0 is very rarely a good estimate. However, a possible explanation might be that 0-imputation provides some auto-correction for the bias, since the 0s now describe the missingness pattern.
This result is in accordance  with what Hu et al. found in~\cite{hu2017strategies} when taking a manual feature design approach to the problem.

Regarding bias correction, it is maybe not so surprising that it does not lead to improved performance on the 0-imputed dataset for any of the two kernels, since  0-imputation in itself seems to have the same effect (ref. previous paragraph). On the other hand, since the performance obtained using the linear kernel combined with mean or LOCF imputation is substantially improved, whereas nothing happens for the GAK kernel, it is difficult to draw conclusions about the effect of bias correction.

We also note that the best imputed data methods all share the same pattern; the performance peaks around postoperative window size 11. The natural behaviour should be that the F1-score should increase as more information is added when the length of the postoperative window is increased. 
However, when the opposite happens, this indicates that all these imputation methods introduce a bias into the data that affects the performance.
Hence, due to the high variance in the performance across different imputation methods and length of postoperative window, and since is difficult to do model selection in the unsupervised setting of our framework, we conclude that the linear kernel and GAK are not suitable for the task under analysis.

Not surprisingly, the time series kernels, TCK, LPS and GAK perform worse than the linear kernel when the postoperative window is short ($< 10$ days). When the MTS are shorter than 10 days the time dependency and time structure is probably not clear enough to be fully utilized by the specialized time series kernels. However, as the size of the postoperative window increases, the F1-score obtained using LPS and TCK increase smoothly, before it stabilizes around a F1-score of 0.80 when the length is around 14-15 days, considerably higher than the F1-score obtained using the GAK and linear kernel.
This behaviour verifies the robustness of these two kernels with respect to varying length of postoperative window and missingness patterns.

We note  that for the TCK and LPS kernels an underlying assumption is that values are MAR, whereas the missingness for EHR-data could be partly due to MNAR. 
However, it has been demonstrated by several authors that a slightly wrong assumption of MAR in many realistic scenarios do not have a big impact in terms of biased predictions~\cite{collins2001comparison, schafer2002missing, hu2017strategies}. In our case we do not know by how much the assumption of MAR is broken. % but we do know that the fraction of missing data is very large. 
However, experiments in~\cite{mikalsen2017time} demonstrated these kernels' (and in particular TCK's) robustness to large fractions of missing data -- also in the case of MNAR data, whereas the imputation methods suffered in cases with much missing data. Even though, the data considered in this paper is completely different, it is interesting to observe a similar behaviour here. 
%Due to ensemble strategy ?

%\textbf{mention somewhere that knn imputation is not used due to large amounts of missing data, MI because unsupervised ?}

%Given that it is difficult to do model selection in the unsupervised setting of our framework, and the unstable performance of the GAK and linear kernel and the manual feature baseline, across different imputation methods and length of postoperative window, in contrast to the stability of the TCK and LPS, we conclude that the proposed method is suitable for the task under analysis. \commentC{This sentence is long, I am not sure about what you want to say}

\subsection{Limitations and further work}
%There are also some limitations with respect to how we dealt with the clinical problem we considered in this paper, namely classification of patients with SSI based on blood test.
Although the results obtained in this paper are promising just based on blood results, previous studies~\cite{soguero2016predicting} have shown that the combination of heterogeneous data sources (e.g. free text, drugs, ICD-9 or vital signs) from the EHR might provide better performance. Including more data, however, comes at the cost of a more complicated and computationally demanding analysis.
In addition, we did not differentiate between deep and superficial SSI in this work. These issues are subject to further work.
Moreover, in this work, we focused on identifying patients with SSI based on postoperative data. In further work we would like to do prediction of SSI based on preoperative data and investigate if we can predict that the patient gets SSI before he or she is diagnosed with the complication, which is a framework that would be very valuable to operationalize in the clinic. 
The framework used in this work is general, and not restricted to SSI, it will therefore be interesting to also see how well it works on detecting other complications or diseases.

We could have benchmarked the proposed method against other kernels as well, but chose to not do that for the clarity of the presentation, and because the results obtained using GAK and the linear kernel indicate that results are more dependent on the imputation method than the type of kernel. We also tried with different clustering algorithms in the last step such as hierarchical clustering and kNN mode seeking ensemble clustering~\cite{NordhaugMyhre2018491}, but initial experiments did not improve performance, and we have therefore, for the conciseness of the presentation, not included them. %\commentF{not sure about justifying ourselves for not having tried all the possible existing approaches. I would rather say that we used what we did because we found it suitable for certain reasons.}

%1.- Assessment of surgical site infection risk factors at Imam Reza hospital, Mashhad, Iran between 2006 and 2011  --> 'The missing data included duration of surgery and pre-operative and post-operative hospital stay in 40$\%$ and 33$\%$ of patients, respectively.'

%2.-  Developing a Risk Stratification Model for Surgical Site Infection after Abdominal Hysterectomy --> 'In order to include all subjects in the analysis, multiple imputation was used to create values for missing body mass index, serum glucose and creatinine.  We used multiple imputation to impute a set of plausible glucose values that represent the uncertainty about the correct value'

%\begin{enumerate}
%\item Number of available data before surgery
%\item We don't when the patient are sufering SSI, did they %annotated that after the surgery? Check
%\item After surgey, many more blood tests available but with not %measure at all.
%\item Say something about labels
%\end{enumerate}

\section{Conclusions}
\label{sec: conclusion}
Hospital acquired infections in general, and surgical site infection in particular,  are major problems at modern hospitals nowadays. To be able to reduce this problem, accurate prediction of SSI is of utmost importance. In this study, we showed that analyzing EHR data as MTS within a kernel framework can be very powerful in that respect.  In particular, the LPS and TCK kernels that explicitly can deal with the missing data, turn out to be more robust and work better than those kernels that require the incomplete data to be pre-processed using some imputation method.

Moreover, because of the two kernels' robustness to hyperparameters, we showed that we can completely unsupervised identify patients with SSI and perform similarly to a supervised baseline, hence alleviating the problem of a time consuming and expensive manual label annotation process, often unfeasible for large datasets. Worth mentioning in that respect, is that in this paper we have also illustrated the power of using only blood tests as the data source, hence also reducing the burden for the patients and data engineers.

\section*{Conflict of interest}
The authors have no conflict of interest related to this work.

\section*{Acknowledgement}
 This work was partially funded by the Norwegian Research Council FRIPRO grant no. 239844 on developing the \emph{Next Generation Learning Machines}.
Cristina Soguero-Ruiz is partially supported by project TEC2016-75361-R from Spanish Government and by project DTS17/00158 from Institute of Health Carlos III (Spain).

The authors would like to thank Kristian Hindberg from UiT The Arctic University of Norway for his assistance on preprocessing and extracting the data from the EHR system. We would also like to thank Rolv-Ole Lindsetmo and Knut Magne Augestad from the University Hospital of North Norway, Fred Godtliebsen from UiT, together with Stein Olav Skrøvseth from the Norwegian Centre for E-health Research for helpful discussions throughout the study and manuscript preparation.

\appendix

\section{TCK}
\label{appendix: TCK}

\subsection*{Notation}

%Since the linear span of the posterior distributions, equipped with ordinary inner product, constitutes a Hilbert space, it follows that the TCK is a kernel where the feature map is explicitly given via the posteriors, \textbf{normalized in the l2-norm}.

The following notation is used.
A  multivariate time series (MTS) $X$ is defined as a (finite) sequence of univariate time series (UTS),
$
X = \{ x_v \in \mathbb{R}^T \: | \: v = 1,2,\dots,V\},
$
where each attribute, $x_v$, is a UTS of length $T$. The number of UTS, $V$, is the \textit{dimension} of $X$. The length $T$ of the UTS $x_v$ is also the length of the MTS $X$. Hence,  a $V$--dimensional MTS, $X$, of length $T$ can be represented as a matrix in $\mathbb{R}^{V \times T}$. 
Given a dataset of $N$ MTS, we denote $X^{(n)}$ the $n$-th MTS. 
An incompletely observed MTS is described by the pair $(X^{(n)}, R^{(n)})$, where $R^{(n)}$ is a binary MTS 
with entry $r_v^{(n)}(t) = 0$ if the realization $x_v^{(n)}(t)$ is missing and $r_v^{(n)}(t) = 1$ if it is observed.

\subsection*{DiagGMM}

To build the TCK kernel matrix, we first fit different diagonal covariance GMM (DiagGMM) to the MTS dataset.
In the DiagGMM one assumes time-dependent means, expressed by $\mu_g = \{ \mu_{gv} \in  \mathbb{R}^T \: | \: v = 1,...,V\}$, where  $\mu_{gv}$ is a UTS, and a time-constant covariance matrix is $\Sigma_g = diag\{\sigma_{g1}^2,...,\sigma_{gV}^2\}$, being $\sigma_{gv}^2$ the variance of attribute $v$.
Moreover, the data is assumed to be \textit{missing at random} (MAR), i.e. the missing elements are only dependent on the observed values.
Under these assumptions, missing data can be analytically integrated away, such that imputation is not needed~\cite{rubin1976inference}, and the pdf for the incompletely observed MTS $(X, R)$ is given by
\begin{equation} \label{eq: p(x) gmm diag}
p(X \: | \: R, \: \Theta ) = \sum_{g=1}^G \theta_g \prod_{v=1}^V \prod_{t=1}^T  \mathcal{N} (x_v(t) \: | \: \mu_{gv}(t), \sigma_{gv})^{r_v(t) }
\end{equation}
The conditional probability of $Z$ given $X$, can be found using Bayes' theorem,
\begin{equation} \label{eq: p(z|x) posterior}
\pi_{g} \equiv P(Z_g = 1 \: | \: X, \: R, \: \Theta )  
= \frac{ \theta_g \prod_{v=1}^V \prod_{t=1}^T  \mathcal{N} \left(x_v(t) \: | \: \mu_{gv}(t), \sigma_{gv}\right)^{r_v(t) }}{\sum_{g=1}^G \theta_g \prod_{v=1}^V \prod_{t=1}^T  \mathcal{N} \left(x_v(t) \: | \: \mu_{gv}(t), \sigma_{gv}\right)^{r_v(t) }}.
\end{equation}
The parameters of the DiagGMM are learned using a maximum a posteriori expectation maximization algorithm, as described in \cite{mikalsen2017time}. 

\subsection*{Ensemble strategy}
To ensure diversity, each GMM model uses a number of components from the interval $[2,C]$, where $C$ is the maximal number of mixture components. For each number of components, we apply $Q$ different  random initial conditions and hyperparameters. We let $\mathcal{Q} = \{ q = (q_1,q_2) \: | \: q_1=1,\dots Q, \: q_2 = 2,\dots, C \} $ be the index set keeping track of initial conditions and hyperparameters ($q_1$), and the number of components ($q_2$).
Moreover, each model is trained on a random subset of MTS, accounting only a random subset of variables $\mathcal{V}$, with cardinality $ |\mathcal{V}| \leq V$, over a randomly chosen time segment $\mathcal{T}, |\mathcal{T}| \leq T$. 
The inner products of the posterior distributions from each mixture component are then added up to build the TCK kernel matrix, according to the ensemble strategy~\cite{ensemble}.
Algorithm~\ref{alg:algorithm} describes the details of the method.

% :::::::::::::::::::::: ALGO TCK in-sample ::::::::::::::::::::::
\begin{algorithm}[t!]
%\footnotesize
\small
\caption{TCK kernel. Training phase.}
\label{alg:algorithm}
\begin{algorithmic}[1]
\Require Training set of MTS $ \{ X^{(n)}  \}_{n=1}^N$ , $Q$ initializations, $C$ maximal number of mixture components.
\State Initialize kernel matrix $K = 0_{N \times N}  $.
\For{$q \in \mathcal{Q}$}
\State Compute posteriors $ \Pi^{(n)}(q) \equiv ( \pi_1^{(n)},\dots,\pi_{q_2}^{(n)} )^T $, by applying maximum a posteriori expectation maximization~\cite{mikalsen2017time} to the DiagGMM with $q_2$ clusters and by randomly selecting,
%{\setstretch{0.4}
\begin{itemize}
\item[i.] hyperparameters $\Omega(q) $,
\item[ii.] a time segment $ \mathcal{T}(q)  $ of length {\footnotesize $T_{min} \leq  |\mathcal{T}(q)| \: \leq \: T_{max}$ },
\item[iii.] attributes $\mathcal{V}(q)$, with cardinality {\footnotesize$V_{min} \leq |\mathcal{V}(q)| \leq V_{max}$},
\item[iv.] a subset of MTS, $\eta(q) $, with {\footnotesize$N_{min} \leq |\eta(q)| \leq N$},
\item[v.] initialization of the mixture parameters $ \Theta(q) $.
\end{itemize}
%}
\State Update kernel matrix, $K_{nm} = K_{nm} + \frac{\Pi^{(n)}(q)^T \Pi^{(m)}(q)}{ \| \Pi^{(n)}(q) \| \| \Pi^{(m)}(q) \| } $.
\EndFor
\Ensure $K$ TCK matrix, time segments $\mathcal{T}(q)  $, subsets of attributes $\mathcal{V}(q)$, subsets of MTS $\eta(q)$, parameters $ \Theta(q)$  and posteriors $\Pi^{(n)}(q) $.
\end{algorithmic}
\end{algorithm}
% :::::::::::::::::::::::::::::::::::::::::::::::::::::::::::::::::

\subsection*{Method details}
Algorithm~\ref{alg:algorithm} describes the details of the method. $\mathcal{Q} = \{ q = (q_1,q_2) \: | \: q_1=1,\dots Q, \: q_2 = 2,\dots, C \} $ is the index set keeping track of initial conditions and hyperparameters ($q_1$), and the number of components ($q_2$).

In order to be able to compute similarities with MTS not available at the training phase, one needs to store the time segments $\mathcal{T}(q)$, subsets of attributes $\mathcal{V}(q)$, DiagGMM parameters $ \Theta(q)$  and posteriors $\mathbf{\Pi}^{(n)}(q)$.
Then, the TCK for such out-of-sample MTS is evaluated according to Algorithm~\ref{alg:algorithm out of sample}.

% :::::::::::::::::::::: ALGO TCK out-sample ::::::::::::::::::::::
\begin{algorithm}[h!]
%\footnotesize
\small
\caption{TCK kernel. Test phase.}
\label{alg:algorithm out of sample}
\begin{algorithmic}[1]
  \Require Test set $\big \{ X^{*(m)} \big \}_{m=1}^M$, time segments $\mathcal{T}(q)$, attributes $\mathcal{V}(q)$,  subsets of MTS $\eta(q)$, parameters $  \Theta(q) $  and posteriors $\Pi^{(n)}(q) $. 
  \State Initialize kernel matrix $K^* = 0_{N \times M} $. 
  \For{$q \in \mathcal{Q}$}
  \State Compute posteriors $\Pi^{*(m)}(q) $, $m=1,\dots,M$ using the mixture parameters $ \Theta(q)$.
  \State Update kernel matrix, $K^*_{nm} = K^*_{nm} + \frac{\Pi^{(n)}(q)^T \Pi^{*(m)}(q)}{ \| \Pi^{(n)}(q) \| \| \Pi^{*(m)}(q) \| } $.
  \EndFor
  \Ensure $K^*$ TCK test kernel matrix.
\end{algorithmic}
\end{algorithm}
% :::::::::::::::::::::::::::::::::::::::::::::::::::::::::::::::::

%\section{Learned pattern similarity}
%\label{appendix: LPS}

%##########################
% LPS
%###########################################################

%probably the lps should be described in a bit more detail.
\if
In the LPS a time series is represented as a matrix of segments. Randomness is injected to the learning process by randomly choosing time segment (column in the matrix) and lag $p$ for each tree in the Random Forest.
For each tree a Bag-of-Words type compressed representation is created from the output of the leaf-nodes. The
final time series representation is created by concatenating the representation obtained from the individual trees and is  in turn used to design the similarity using a histogram intersection kernel. 

Given two multivariate time series $X^{(n)}$ and $X^{(m)}$, a formal expression for the LPS-kernel is 
\begin{equation} \label{eq: LPS}
    K(X^{(n)},X^{(m)}) = \frac{1}{R J} \sum\limits_{k=1}^{R J} \min (h^n_k, h^m_k),
\end{equation}
where $h^n_k$ is the $k$th entry of the concatenated Bag-of-Words representation $H(X^{(n)})$.  More precisely, $H(X^{(n)})$ is a concatenation of $R$-dimensional frequency vectors of instances in the terminal nodes from all trees.

Each tree generates a representation and the
final time series representation is obtained via concatenation. For simplicity, assume
that all trees contain the same number of terminal nodes R. The general case is easily
handled. Let H j (x n ) denote the R-dimensional frequency vector of instances in the
terminal nodes from tree g j for time series x n . We concatenate the frequency vectors
over the trees to obtain the final representation of each time series, denoted as H (x n ),
of length R  J (and modified obviously for non-constant R). Our representation
summarizes the patterns in the time series based on the terminal node distribution of
the instances over the trees.
\fi

\if
%from the great time series bake-off:
For each location, the subseries in the original data are concatenated to form a
new attribute. The internal model selects a random attribute as the response
variable then constructs a regression tree. A collection of these regression trees
are processed to form a new set of instances based on the counts of the number

of subseries at each leaf node of each tree. Algorithm 8 describes the process.
LPS can be summarised as follows:
Stage 1: Construct an ensemble of r regression trees.
1. Randomly select a segment length l (line 3)
2. Select w segments of length l from each series storing the locations in matrix
A (line 4).
3. Select w segments of length l from each di↵erence series storing the locations
in matrix B (line 5).
4. Generate the n · l cases each with 2w attributes (line 6).
5. Choose a random column from W as the response variable then build a
random regression tree (i.e. a tree that only considers one randomly selected
attribute at each level) with maximum depth of d (line 7).
Stage 2: Form a count distribution over each tree’s leaf node.
1. For each case x in the original data, get the number of rows of W that
reside in each leaf node for all cases originating from x.
2. Concatenate these counts to form a new instance. Thus if every tree had t
terminal nodes, the new case would have r · t features. In reality, each tree
will have a di↵erent number of terminal nodes.
Classification of new cases is based on a 1-nearest neighbour classification on
these concatenated leaf node counts.
\fi

\section*{References}
\bibliographystyle{elsarticle-num}
\bibliography{bibliography}

\end{document}